\documentclass[conference]{IEEEtran}
\IEEEoverridecommandlockouts
\usepackage[T1]{fontenc}
\usepackage{amsmath,amssymb}
\usepackage{graphicx,booktabs,array,tabularx}
\usepackage{cite,url,xcolor,listings}
\usepackage{capt-of}
\usepackage{needspace}
\usepackage[most]{tcolorbox}
\usepackage[hidelinks]{hyperref}
\hypersetup{pdftitle={When News Gets Old: Portfolio Allocation with Reinforcement Learning and Large Language Models},pdfauthor={Bolin Song; Surui Xiang}}
\title{When News Gets Old: Portfolio Allocation\\with Reinforcement Learning\\and Large Language Models}
\author{\IEEEauthorblockN{Bolin Song}
\IEEEauthorblockA{Cornell University\\bs833@cornell.edu}
\and
\IEEEauthorblockN{Surui Xiang}
\IEEEauthorblockA{United Nations\\surui.xiang@un.org}}
\begin{document}
\raggedbottom
\maketitle
\begin{abstract}
Financial news can influence investment decisions long after publication.
Timely news may reach an investor before prices fully adjust, whereas older
news may already be reflected in prices. We examine two ways to handle aging
signals: retain them and give a portfolio policy timing information, or
filter them out before portfolio selection. A fixed large language model
(LLM) extracts signals from public SEC disclosures. We compare these
treatments in reinforcement learning (RL) policies and LLMs that choose
portfolios directly. We evaluate returns after trading costs in historical
tests over 2023--2025. RL choices respond to timing information in a controlled
test, but historical returns show no consistent benefit from adding it. For the LLM, filtering
old news produced the highest average net return, reaching 18.43\% per year
when news delivery was delayed by four weeks. This exceeded the returns of
all RL strategies that used news.

\end{abstract}

\section{Introduction}
A company reports stronger sales, and a large language model (LLM) assigns
the report a positive signal. A month later, should that signal still
influence the portfolio in the same way? Prices may already reflect the
news, and the company's outlook may have changed. The report still describes
a favorable development, but its usefulness for a new trade is less certain.

Reinforcement learning (RL) has been used to learn trading decisions that
account for existing positions and transaction costs~\cite{moody2001,jiang2017}.
LLM-assisted systems supply textual signals to RL~\cite{benhenda2025}, while
other LLM agents use financial memory to make decisions directly~\cite{yu2023}.
The value of acting on a signal depends on how long it remains useful and
on trading costs~\cite{garleanu2013,garleanu2016}. Tetlock~\cite{tetlock2011}
examines stale information through repeated news content. We focus on a
practical choice as a report grows older: retain its signal and tell the
portfolio policy about its timing, or remove the signal before the policy
makes a decision.

We compare these approaches in a weekly allocation task on US stocks. An
LLM reads periodic SEC disclosures and assigns positive, neutral, or negative signals
based on the company's business performance or outlook. Either an RL policy or an LLM then uses these
signals, market data, and current holdings to choose a portfolio. Both choose
from the same predefined portfolios and can keep their existing holdings.
We compare three input settings: signals without timing information, signals
with timing information, and signals filtered out once the source report
is more than eight weeks old.

We measure returns after trading costs over 2023--2025, making reports
available without added delay or holding them back for one or four weeks.
Separately, we make already available reports appear older while keeping
market data, holdings, and extracted signals unchanged. This second test
checks whether news age alone changes the selected portfolio.

Providing timing information changes RL choices in the controlled test
but does not consistently improve historical net returns.
For the LLM portfolio policy, filtering
old signals yields the highest average net return. With a four-week delivery
delay, it earns 18.43\% per year after costs, exceeding all three RL policies
that use news. A simple age filter can therefore offer a useful alternative
to retaining older signals and asking the portfolio model to interpret their
timing.

\begin{figure*}[t]
\centering
\includegraphics[width=\textwidth]{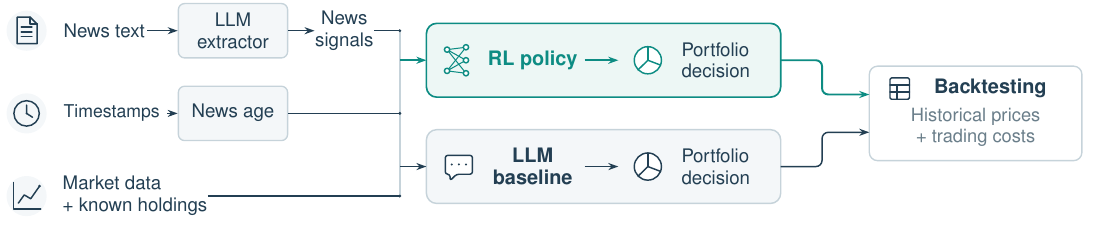}
\caption{\textbf{News Signals and Portfolio Decisions.} An LLM extracts news
signals, while report dates determine news age. The RL policy and direct
LLM baseline choose portfolios independently, with separate holdings under
common backtesting rules. Table~\ref{tab:methods} specifies each method's inputs.}
\label{fig:information}
\end{figure*}

\section{Decision Framework}
\subsection{Observations and News Timing}
We use an LLM to extract positive, neutral, or negative signals from
companies' periodic SEC disclosures. Each week, RL or an LLM combines
these signals with market data and holdings to choose from the same target
portfolios or keep its holdings. Signals persist until newer reports arrive,
unless filtered.

Consider $N$ assets and weekly decisions indexed by $t$. The policy observes
\begin{equation}
o_t=(\bar{\mathbf w}_t,\mathbf m_t,\mathbf z_t,
\boldsymbol\tau_t,\mathbf g_t),
\label{eq:observation}
\end{equation}
where $\bar{\mathbf w}_t$ contains portfolio weights known at the decision
time, $\mathbf m_t$ contains past returns and volatility, and
$z_{i,t}\in\{-1,0,1\}$ is a negative, neutral, or positive signal about the
company's business performance or outlook.

News age measures the number of weeks since a report was published. Time
since arrival measures how long the report has been available to the policy.
We refer to these two inputs together as timing information. For example,
a report published four weeks ago but arriving today has a news age of four
weeks and a time since arrival of zero. These quantities form
$\boldsymbol\tau_t$ and $\mathbf g_t$, respectively.

Let $b_t$ be decision time in weeks, $u_{i,t}$ the publication time of the
latest visible report, and $v_{i,t}=u_{i,t}+\delta$ its arrival under delay
$\delta$. Then
\begin{equation}
\tau_{i,t}=b_t-u_{i,t},\qquad g_{i,t}=b_t-v_{i,t},\qquad v_{i,t}\leq b_t.
\label{eq:time}
\end{equation}
A fixed LLM reads each report once and returns a report signal with a short
quotation from the report. Its
input contains only the report. All policies share the stored signals.
A zero signal can mean a neutral report, no available report, or an output
that failed a format or evidence check. Appendix~\ref{app:prompts} gives the prompts and a recorded
report-to-portfolio example.

We compare each news treatment against \emph{News}, which uses market data,
holdings, and report signals without timing information. \emph{News + Age}
adds the two timing inputs. \emph{Filtered News} sets a report's signal to
zero once it is more than eight weeks old, without supplying timing inputs.
The stock remains available for investment. The filtered signal is
\begin{equation}
\widetilde z_{i,t}=z_{i,t}\mathbf1\{\tau_{i,t}\leq h\},
\label{eq:filter}
\end{equation}
where the common age limit is $h=8$ weeks for RL and LLM portfolio policies.
Table~\ref{tab:methods} includes the market-only and passive benchmarks.

\begin{table*}[t]
\centering\small
\caption{Decision inputs and comparison methods.}
\label{tab:methods}
\begingroup
\setlength{\tabcolsep}{2.3pt}
\renewcommand{\arraystretch}{1.10}
\begin{tabularx}{\textwidth}{@{}p{.050\textwidth}p{.245\textwidth}*{4}{>{\centering\arraybackslash}X}@{}}
\toprule
Policy & Input setting & \shortstack{Market data\\\& holdings} & \shortstack{News\\signals} & \shortstack{News age\\\& time since arrival} & \shortstack{Decision\\maker} \\
\midrule
\multicolumn{2}{@{}l}{Equal Allocation} & --- & --- & --- & Fixed weights \\
\multicolumn{2}{@{}l}{Buy and Hold} & --- & --- & --- & Fixed positions \\
\midrule[.25pt]
RL & Market Data & $\checkmark$ & --- & --- & RL \\
 & Market Data + News & $\checkmark$ & $\checkmark$ & --- & RL \\
 & Market Data + News + News Age & $\checkmark$ & $\checkmark$ & $\checkmark$ & RL \\
 & Market Data + Filtered News & $\checkmark$ & $\checkmark^{*}$ & --- & RL \\
\midrule[.25pt]
LLM & Market Data & $\checkmark$ & --- & --- & LLM \\
 & Market Data + News & $\checkmark$ & $\checkmark$ & --- & LLM \\
 & Market Data + News + News Age & $\checkmark$ & $\checkmark$ & $\checkmark$ & LLM \\
 & Market Data + Filtered News & $\checkmark$ & $\checkmark^{*}$ & --- & LLM \\
\bottomrule
\multicolumn{6}{@{}p{0.98\textwidth}@{}}{$^{*}$Reports older than eight weeks have their signals set to zero; numerical ages are not passed to the filtered policy.} \\
\end{tabularx}
\endgroup

\end{table*}

\subsection{Actions, Reward, and Learning}
A target portfolio specifies how much capital to allocate to each stock.
For example, one target assigns 40\% to one stock and spreads the remaining
60\% equally across the remaining stocks. Each week, the policy selects a target
from the predefined list or keeps its current holdings. Rebalancing to the
selected target incurs trading costs. HOLD keeps the existing shares.
All strategies are fully invested and use the same execution schedule.
Equal Allocation restores $q_i=1/N$ each week. Buy and Hold starts at the
same weights and preserves its shares, allowing weights to drift.

Let $V_t$ be wealth before trading, with subsequent wealth $V_{t+1}$ already
reflecting trading costs. The reward and objective are
\begin{equation}
r_{t+1}=\kappa\log\frac{V_{t+1}}{V_t},\qquad
\max_\pi\;\mathbb E_\pi\!\left[\sum_{t=0}^{T-1}\gamma^t r_{t+1}\right].
\label{eq:objective}
\end{equation}
Here $T$ is the horizon, $\gamma$ discounts later rewards, and $\kappa>0$
sets reward scale.
Appendix~\ref{app:implementation} gives the execution and cost accounting.

Each RL information setting has a separately trained Double DQN
policy~\cite{vanhasselt2016}, with the same architecture and training budget.
During evaluation, the
trained network chooses the action with the highest predicted value,
without random exploration or further training.
LLM portfolio policies receive the corresponding numerical fields and choose
from the same list of portfolios. Assets are represented by anonymous indices;
company names, dates, and raw reports are omitted. News and filtered-news prompts are
identical; only the supplied signals change. Invalid action outputs map to
HOLD. Each policy maintains its own evolving holdings.

\begin{table*}[t]
\centering\fontsize{8.5}{10}\selectfont
\setlength{\tabcolsep}{2.3pt}
\renewcommand{\arraystretch}{1.12}
\caption{Historical returns and trading outcomes.}
\label{tab:outcomes}
\begin{tabularx}{\textwidth}{@{}p{.050\textwidth}p{.245\textwidth}*{3}{>{\centering\arraybackslash}p{.065\textwidth}}*{4}{>{\centering\arraybackslash}X}@{}}
\toprule
\multicolumn{2}{c}{} & \multicolumn{3}{c}{\shortstack{Annualized return\\after costs}} & \multicolumn{4}{c}{Results with a four-week news delay} \\
\cmidrule(lr){3-5}\cmidrule(lr){6-9}
Policy & Input setting & 0 weeks & 1 week & 4 weeks & Sharpe ratio & Maximum drawdown & Annual trading volume ($\times$) & Total trading costs \\
\midrule
\multicolumn{2}{@{}l}{Equal Allocation} & 17.63\% & 17.63\% & 17.63\% & 1.13 & 10.44\% & 1.09 & 0.12\% \\
\multicolumn{2}{@{}l}{Buy and Hold} & 17.19\% & 17.19\% & 17.19\% & 1.13 & 10.45\% & 0.00 & 0.00\% \\
\midrule[.3pt]
RL & Market Data & 19.07\% & 19.07\% & 19.07\% & 1.00 & 12.99\% & 25.12 & 2.69\% \\
 & Market Data + News & 16.22\% & 16.95\% & 17.81\% & 0.96 & 12.49\% & 23.72 & 2.55\% \\
 & Market Data + News + News Age & 15.41\% & 15.11\% & 18.02\% & 0.98 & 12.88\% & 24.75 & 2.64\% \\
 & Market Data + Filtered News & 18.54\% & 19.20\% & 18.03\% & 0.97 & 13.03\% & 25.40 & 2.73\% \\
\midrule[.3pt]
LLM & Market Data & 18.02\% & 18.02\% & 18.02\% & 1.09 & 11.07\% & 20.97 & 2.25\% \\
 & Market Data + News & 17.60\% & 16.19\% & 16.02\% & 0.98 & 12.01\% & 18.87 & 2.02\% \\
 & Market Data + News + News Age & 16.73\% & 13.26\% & 17.17\% & 1.07 & 11.31\% & 13.01 & 1.36\% \\
 & Market Data + Filtered News & 20.16\% & 18.35\% & 18.43\% & 1.12 & 10.83\% & 19.03 & 2.05\% \\
\bottomrule
\end{tabularx}

\par\smallskip
\begin{minipage}{\textwidth}\footnotesize
Metrics are computed within each annual test and averaged equally; RL also
averages separately trained policies. The Sharpe ratio uses a zero risk-free
rate. Maximum drawdown is the largest peak-to-trough decline. Annual trading
volume sums buys and sells relative to portfolio value. Total costs are fees
relative to each year's initial capital, then averaged. All methods use the same equal-weight starting portfolio with no initial
trading fee and no forced terminal sale.
\end{minipage}
\end{table*}

\section{Experimental Design}
\subsection{Historical Comparison}
We evaluate weekly portfolio allocation for ten US stocks over 2023--2025,
using historical prices and signals from periodic SEC disclosures. RL
training and validation use data from years before each annual test. Every
method begins each test year with an equal-weight portfolio.

Each week, the portfolio is chosen using information available at that
week's last close. Trades execute at the following week's first open and
cost 10 basis points per dollar bought or sold. We test report delivery
delays of zero, one, and four weeks; report text and publication dates
remain unchanged. All methods use the same underlying data and trading
rules. Each RL policy is trained using all three delay settings. Stock
selection and training details are given in Appendix~\ref{app:implementation}.

GPT-4.1 mini extracts the report signals and selects portfolios for the
LLM results in Table~\ref{tab:outcomes}. At a four-week delivery delay, we
also compare Llama 3.3 and GPT-6 Astra as portfolio decision makers using
the same signals. Model settings are given in Appendix~\ref{app:prompts}.

\subsection{Decision Changes and Net Returns}
Historical tests compare investment returns when reports arrive with an
additional delay of zero, one, or four weeks.

A separate test asks whether older reports change RL's portfolio choice.
Reports, market data, holdings, and original signals stay fixed; news age
and time since arrival increase equally, leaving delivery delay unchanged.
Filtered News re-applies the eight-week cutoff, potentially zeroing more
signals. We measure changes in portfolio choice, not returns from those
changes. Let $o_t^{+\ell}$ denote the input after adding $\ell$ weeks to both
timing inputs.
\begin{equation}
\rho(\ell)=\frac{1}{|\mathcal B|}\sum_{t\in\mathcal B}
\mathbf1\{\pi(o_t^{+\ell})\neq\pi(o_t)\},
\label{eq:response}
\end{equation}
computed over fixed validation observations $\mathcal B$ for each trained
policy. News-only inputs do not change.

Investment value is evaluated on historical prices using net annualized
return, Sharpe ratio, maximum drawdown, trading volume, and fees.
Within each decision method, we subtract the annualized return of News
from those of News + Age and Filtered News, reporting these differences in \%.
Metrics are averaged equally across annual tests.

Following uncertainty-aware RL evaluation~\cite{henderson2018,agarwal2021},
we report training variation separately from uncertainty in return differences.
Paired time-block intervals resample the same weeks across methods and trained
policies. Appendix~\ref{app:diagnostics} gives annual results, training repetitions,
and resampling details. Supporting checks vary fees and examine signal
persistence in a synthetic market (Appendix~\ref{app:implementation}).

\begin{figure*}[t]
\centering
\newsavebox{\resultBehaviorImage}
\newsavebox{\resultReturnImage}
\sbox{\resultBehaviorImage}{\includegraphics[width=0.485\textwidth]{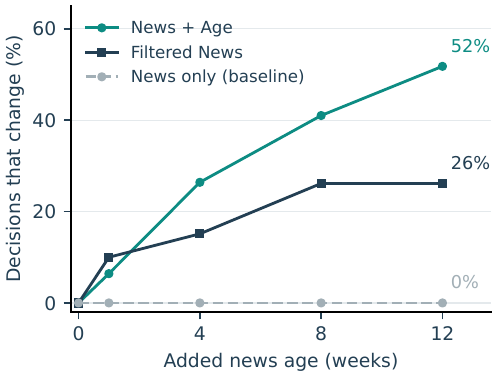}}
\sbox{\resultReturnImage}{\includegraphics[width=0.485\textwidth]{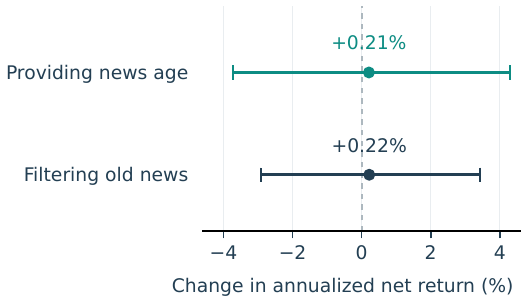}}
\newlength{\resultImageHeight}
\setlength{\resultImageHeight}{\dimexpr\ht\resultBehaviorImage+\dp\resultBehaviorImage\relax}
\begin{minipage}[t]{0.485\textwidth}
\vspace{0pt}
\centering
\begin{minipage}[c][\resultImageHeight][c]{\linewidth}
\centering\usebox{\resultBehaviorImage}
\end{minipage}
\caption{Portfolio decisions as reports age. Market features, holdings, and
original LLM signals stay fixed while publication and arrival times move backward
together. Filtering reapplies its cutoff. Lines average trained policies;
Appendix~\ref{app:diagnostics} reports variation across training runs.}
\label{fig:behavior}
\end{minipage}\hfill
\begin{minipage}[t]{0.485\textwidth}
\vspace{0pt}
\centering
\begin{minipage}[c][\resultImageHeight][c]{\linewidth}
\centering\usebox{\resultReturnImage}
\end{minipage}
\caption{Changes in RL annualized net return relative to Market Data + News
at a four-week delivery delay. Providing news age adds both timing inputs;
filtering removes signals older than eight weeks. Dots are paired mean
differences; bars are 95\% time-block intervals
(Appendix~\ref{app:diagnostics}).}
\label{fig:returns}
\end{minipage}
\end{figure*}

\Needspace{9\baselineskip}
\section{Results}
\label{sec:results}
\subsection{LLM Portfolio Returns}
Filtering old news gives the LLM its highest average net return
(Table~\ref{tab:outcomes}). With a four-week delivery delay, removing signals
from reports more than eight weeks old produces an average annualized net
return of 18.43\%. Keeping the signals produces 16.02\% without timing
information and 17.17\% with timing information. The LLM with filtering also
earns a higher return than all three RL policies that use news.
RL with Market Data returns 19.07\%, above the LLM's 18.43\% with filtering.
News does not always improve on market data alone, and its treatment matters.

Additional LLM comparisons show that the effect of adding timing information
varies by model and year (Appendix~\ref{app:diagnostics}).

\Needspace{9\baselineskip}
\subsection{RL Responses to News Age}
RL changes its portfolio choices in response to news age
(Fig.~\ref{fig:behavior}). When the same reports are made twelve weeks older,
RL with News + Age selects a different portfolio in 52\% of cases, averaged
across trained policies on 39 fixed validation observations. Market data,
holdings, and the original report signals remain
unchanged in this test.

RL with Filtered News changes its choice in 26\% of cases under the same
twelve-week increase. Here, making reports older causes more signals to
cross the eight-week cutoff and be removed. Its choices stop changing once
no additional signals are removed. RL with News keeps the same choices
throughout because its inputs do not change.
Decision changes show sensitivity to timing information, without establishing
that RL systematically reduces the influence of older signals.

\subsection{RL Portfolio Returns}
In the historical backtests, adding timing information does not consistently
improve RL net returns. It gives RL lower average returns than
News at delivery delays of zero and one week. With a four-week delay, the
average annualized net return increases only from 17.81\% for News to
18.02\% for News + Age. Filtered News produces a similar return of 18.03\%.
The 95\% confidence intervals for both gains over News include zero
(Fig.~\ref{fig:returns}).

At the four-week delay, filtering removes roughly 70\% of nonzero news
signals. Keeping these older signals and giving RL timing information
offers no clear return advantage over filtering them out.

\Needspace{7\baselineskip}
\subsection{Trading Costs}
Higher trading fees substantially reduce the return of RL with News + Age.
At a four-week delivery delay, we increase the fee from 10 to 30 basis
points per dollar traded while keeping the portfolio decisions fixed. Its
average annualized net return falls from 18.02\% to 12.32\%. Buy and Hold
remains at 17.19\% because it makes no trades after the initial allocation.

\Needspace{14\baselineskip}
\section{Conclusion}
We compared retaining older signals with timing information and filtering
them before portfolio selection, using RL and LLM policies on US stocks
over 2023--2025. All policies shared signals extracted from SEC disclosures.
RL choices responded to timing information in the controlled test, while
historical tests showed no consistent net return improvement from adding it.
The LLM policy performed best when old signals were filtered out. With a
four-week delivery delay, it earned an average annualized net return of
18.43\%, exceeding all three RL policies that used news.

These results highlight signal filtering as a practical part of LLM-based
portfolio design. Retaining older signals and supplying their timing was
less effective for the LLM policy than simply removing them. For RL, the
findings show that responding to news timing and improving investment
performance are distinct outcomes that should be evaluated together.

\bibliographystyle{IEEEtran}
\bibliography{references}
\appendices
\raggedbottom
\section{Annual Results and Additional Diagnostics}
\label{app:diagnostics}
\subsection{Uncertainty and Annual Variation}
Table~\ref{tab:paired} gives the paired intervals for the average return
differences in Fig.~\ref{fig:returns}. For each bootstrap replicate, we sample circular
blocks of eight consecutive weeks separately within each year. A year's
sampled indices are shared by all compared methods and training seeds.
Each trained policy's sampled path is compounded and annualized before
averaging across policies and then equally across years. Percentile intervals use 10,000 replicates, conditional on the recorded
policies and historical periods. Annual segments are resampled independently,
which neglects dependence across year boundaries. Training seeds quantify
variation across training runs and are held together during time resampling.

\begin{center}
\begin{minipage}{\columnwidth}
\centering\footnotesize
\captionof{table}{Paired return differences at four-week delay.}
\label{tab:paired}
\setlength{\tabcolsep}{5pt}
\begin{tabular}{@{}lrr@{}}
\toprule
Comparison & Difference (\%) & 95\% interval (\%) \\
\midrule
News + Age $-$ News & $+0.21$ & $[-3.73, 4.29]$ \\
Filtered News $-$ News & $+0.22$ & $[-2.93, 3.44]$ \\
\bottomrule
\end{tabular}

\end{minipage}
\end{center}

Table~\ref{tab:annual} gives the annual paired contrasts. None of the
36 primary annual timing/filter comparisons has an eight-week-block 95\%
interval excluding zero. Four-week blocks give the same count; with
13-week blocks, only mini filtering at zero delay in 2025 excludes zero.
These are unadjusted exploratory intervals. Variation across training runs is reported
separately: the four-week RL timing contrast spans positive and negative
values across training runs in every year.

\subsection{LLM Portfolio Policies}
The direct-decision comparison keeps the mini-extracted signals fixed and
changes only the model selecting a portfolio. Table~\ref{tab:models}
shows the News + Age versus News return contrasts at four-week delay. All nine
corresponding paired intervals contain zero. The effect of timing varies with both the model and the annual period.

\begin{center}
\begin{minipage}{\columnwidth}
\centering\footnotesize
\setlength{\tabcolsep}{7pt}
\captionof{table}{Timing effects in LLM portfolio policies.}
\label{tab:models}
\begin{tabular}{@{}lrrr@{}}
\toprule
& \multicolumn{3}{c}{Return change (\%)} \\
\cmidrule(l){2-4}
Model & 2023 & 2024 & 2025 \\
\midrule
GPT-4.1 mini & $-3.42$ & $+6.40$ & $+0.48$ \\
Llama 3.3 & $+0.72$ & $+0.54$ & $-1.76$ \\
GPT-6 Astra & $+5.35$ & $+0.50$ & $-4.36$ \\
\bottomrule
\end{tabular}
\end{minipage}
\end{center}

\subsection{Variation in the Behavioral Response}
At twelve weeks of added age, the action-change rates across trained policies
range from 25.6\% to 74.4\% for News + Age and from 7.7\% to 41.0\% for
Filtered News. Figure~\ref{fig:behavior} displays their respective means.

For the same fixed inputs as Fig.~\ref{fig:behavior}, we also measure
$\sum_i z_i(q_i^{\mathrm{older}}-q_i^{\mathrm{base}})$. A positive value
means more weight on favorable signals or less on adverse ones. Unlike
an action-change count, this describes the direction of the allocation
response. The mean does not decline steadily as reports age: some trained
policies increase signal-aligned allocation, while others decrease it.

\begin{table*}[t]
\centering\footnotesize
\caption{Annual return differences at four-week delay, with paired 95\% intervals (\%).}
\label{tab:annual}
\begin{tabular*}{\textwidth}{@{\extracolsep{\fill}}lcccc@{}}
\toprule
& \multicolumn{2}{c}{RL} & \multicolumn{2}{c}{Direct LLM} \\
\cmidrule(lr){2-3}\cmidrule(lr){4-5}
Year & News + Age $-$ News & Filtered News $-$ News & News + Age $-$ News & Filtered News $-$ News \\
\midrule
2023 & $+1.18\;[-4.39, 6.71]$ & $-1.41\;[-5.15, 2.08]$ & $-3.42\;[-17.35, 10.18]$ & $+4.34\;[-4.90, 18.77]$ \\
2024 & $+0.51\;[-3.86, 4.82]$ & $+1.44\;[-4.84, 8.35]$ & $+6.40\;[-7.90, 20.99]$ & $-3.93\;[-24.10, 13.41]$ \\
2025 & $-1.06\;[-10.63, 8.98]$ & $+0.63\;[-5.38, 6.12]$ & $+0.48\;[-8.78, 8.43]$ & $+6.83\;[-6.60, 20.02]$ \\
\bottomrule
\end{tabular*}

\par\smallskip
\begin{minipage}{\textwidth}\footnotesize
Each entry subtracts the News policy's annualized return. Intervals use
paired eight-week time blocks. The direct LLM comparison uses the fixed mini baseline.
\end{minipage}
\end{table*}

\section{Implementation Details}
\label{app:implementation}

\subsection{Self-Financing Execution}
Decision weights $\bar{\mathbf w}_t$ use known closing prices. At the next
execution open, prices may have moved and actual weights are
$\widetilde{\mathbf w}_t$. For a target $\mathbf q$, the fraction of pretrade
wealth left after paying fees solves
\begin{equation}
\eta_t=1-c\sum_i|\eta_t q_i-\widetilde w_{i,t}|.
\label{eq:cost}
\end{equation}
Here $c$ is the fee per dollar bought or sold, including the cash needed to
pay the fee. With asset gross returns $G_{i,t+1}$ until the next execution,
\begin{equation}
V_{t+1}=V_t\eta_t\sum_i q_iG_{i,t+1}.
\label{eq:wealth}
\end{equation}
For HOLD, $\mathbf q_t=\widetilde{\mathbf w}_t$ and $\eta_t=1$; existing
shares evolve with prices without a trade. Fees use actual
execution holdings, while policy inputs use only decision-time information.

\subsection{Annual Folds and Training}
The expanding historical folds below use execution-year cohorts. Counts
exclude one boundary-crossing return from each training and validation
block. A retained return must end no later than the next block's first
decision cutoff. Each annual test starts with an equal-weight portfolio and no initial trading fee. The evaluation is retrospective: 2023 had served as validation in a prior
analysis, and 2024--2025 had already been inspected. Stocks, prompts, budgets,
and the filter were fixed before training the annual policies.

\begin{center}
\footnotesize
\setlength{\tabcolsep}{4pt}
\begin{tabular}{@{}llll@{}}
\toprule
Test year & Training (weeks) & Validation & Test weeks \\
\midrule
2023 & 2020--2021 (103) & 2022 (51) & 52 \\
2024 & 2020--2022 (155) & 2023 (51) & 53 \\
2025 & 2020--2023 (207) & 2024 (52) & 52 \\
\bottomrule
\end{tabular}
\end{center}
Each of the four RL methods is trained anew in every fold with seeds
$11,22,33,44,55,66,77,88,99,110$: 120 trained policies. The final fixed-budget
checkpoint is evaluated greedily. Reserved validation blocks do not select
hyperparameters, seeds, filtering thresholds, or checkpoints. Training
samples 52-week windows with 26 weeks of prior feature history and mixes
delays of 0, 1, and 4 weeks; the folds have 52, 104, and 156 eligible window
starts, respectively. Overlapping training windows reuse observations from the same price history.

\begin{center}
\footnotesize
\setlength{\tabcolsep}{3pt}
\begin{tabular}{@{}p{.43\columnwidth}p{.51\columnwidth}@{}}
\toprule
Training parameter & Fixed value \\
\midrule
Transitions per training run; parallel environments & 500,000; 32 \\
Hidden layers; activation & $64,64$; ReLU \\
Optimizer; learning rate & Adam; $3\times10^{-4}$ \\
Discount; reward multiplier & $0.99$; $100$ \\
Loss; gradient-norm limit & Smooth L1; $10$ \\
Replay capacity; minibatch & 100,000; 256 \\
Replay warmup & 2,048 transitions \\
Gradient updates & One per vector-environment step \\
Target-network hard copy & Every 100 optimizer updates \\
Exploration & $\epsilon:1\to0.05$ over 400,000 transitions, then $0.05$ \\
\bottomrule
\end{tabular}
\end{center}

For a replayed transition, the Double DQN target is
\begin{equation}
y_t=r_{t+1}+\gamma(1-d_t)
 Q_{\bar\theta}\!\left(o_{t+1},\arg\max_a Q_\theta(o_{t+1},a)\right),
\label{eq:dqn}
\end{equation}
where $d_t$ indicates termination, $\theta$ is the online-network parameter
vector, and $\bar\theta$ denotes its periodically copied target. Training
minimizes the Huber (smooth L1) loss between $y_t$ and $Q_\theta(o_t,a_t)$.

\subsection{Inputs and Portfolio Choices}
Market features are cumulative close returns over 1, 4, and 12 weeks,
and the sample standard deviation of the last 12 weekly close returns.
Historical RL networks retain all 80 feature slots; unavailable information
is zero-masked. Returns are divided by $0.05$, volatility by $0.02$, and
both timing inputs by 12; weights and signals are unscaled. Before capping,
$\tau_{i,t}-g_{i,t}=\delta$ for the same visible report. Timing inputs are
capped at 104 weeks; a missing report has signal zero and both timing inputs
104 before method masking. Direct LLMs receive the corresponding unscaled fields.
Filtering sets a signal to zero when its \emph{raw news age}
is strictly greater than eight weeks; neither timing input is passed
to the filtered decision maker.

The historical menu has 57 actions: one equal allocation; one target per
asset with weight $0.4$ and the remaining $0.6$ shared equally among other
assets; 45 targets with two weights $0.3$ and the remaining $0.4$ shared
equally among other assets; and HOLD. The four-asset simulator
has 12: equal allocation, four permutations of $(.55,.15,.15,.15)$, six of
$(.4,.4,.1,.1)$, and HOLD. HOLD retains existing holdings. All target
portfolios are long-only and fully invested.

\Needspace{8\baselineskip}
\subsection{Supporting Simulator and Public Data}
For asset $i$, the hidden state $s_{i,t}\in\{-1,+1\}$ remains unchanged
with probability $p\in\{.85,.97\}$. Weekly gross return $G$ follows
\begin{equation*}
\log G_{i,t+1}=0.0005+0.003s_{i,t}
-\tfrac12(0.02)^2+0.02\varepsilon_{i,t+1},
\end{equation*}
where Gaussian shocks have unit variance and cross-asset correlation
$0.25$. State correlation after $\ell$ weeks is $(2p-1)^\ell$, giving a
controlled persistence mechanism. A report is generated with probability $0.25$ per asset-week
(and at the first warmup step); its semantic label is neutral with
probability $0.20$, otherwise matching the hidden state with probability
$0.80$. Policies observe the stored signals, never that state or label.
Each path has 26 warmup and 104 evaluation weeks. There are 50 validation
and 100 test paths per persistence value, paired across methods and
delays $0,1,4$, with separate split randomness. The 600 synthetic reports are divided into 300 training, 100 validation,
and 200 test texts from disjoint expression-template families. Their semantic
labels describe the report, independently of whether its noisy account matches
the hidden state. Reuse retains document IDs. The simulator uses three training seeds $(11,22,33)$, 32 inputs, 12 outputs,
and the training settings above.

At four-week delay, adding timing to news changes simulated annualized returns
by $+0.35$\% in the faster-changing market and $-0.32$\% in the slower one;
the path-paired 95\% intervals are $[-0.09,0.79]$\% and $[-0.71,0.08]$\%.
Replacing stored signals with the reports' true semantic labels in the first
setting changes the mean return of the fixed RL policy with News + Age by
$-0.02$\%
($[-0.16,0.13]$\%). The replacement labels describe report meaning; the trained policy is held fixed.

The historical stock universe is: AAPL, AXP, BA, CAT, CSCO, CVX, HD, JNJ, KO, VZ.
Selection starts from 30 publicly reported 2019 DJIA ETF holdings~\cite{spdr2019holdings} and 25
issuers with one 10-K and three 10-Q filings in 2019: choose the first
alphabetical eligible ticker in each of eight declared business groups,
then fill alphabetically to the required universe size. Later returns or coverage do not select
or replace stocks. The corpus has 277 SEC excerpts of 800--1,000
words, published from 2019 through 2025. The 2023--2025 cohorts contain 119 reports.
Reports published before a test year remain available until superseded.

The extractor correctly classifies 191 of 200 held-out synthetic reports:
95.5\% accuracy and macro-F1 0.956. For real text, 228 of 277 outputs (82.3\%)
meet the format and verbatim-evidence requirements; 48 rejected outputs and
one request failure become zero signals. The SEC rate measures format and evidence compliance; semantic accuracy
cannot be assessed without human labels for these excerpts.

\subsection{Behavior and Cost Checks}
Behavior probes use every fourth eligible validation week in the 2024 fold,
with all three delays: 39 fixed observations. All already-visible report
publication and arrival times move backward together by 0, 1, 4, 8, or 12
weeks. Holdings follow the equal-allocation reference at known closing prices.
Missing-report sentinels are preserved, and no selected report crosses the
age cap. The ten trained policies use identical observation sets.

The annual cost check replays all ten main methods at four-week news delay,
at 0, 10, and 30 basis points per dollar bought or sold. Each saved action is
held fixed while holdings and fees are recomputed. At 10 basis points, replay reproduces the recorded weekly paths and metrics.
Policies and saved actions are held fixed at the alternative fee levels.

\subsection{Prices and Statistical Summaries}
Daily prices cover 2019 through early January 2026. Adjusted execution open
is vendor open multiplied by adjusted-close/close~\cite{yahooAdjustedClose}.
Adjusted prices provide a current-vintage proxy for total returns. SEC acceptance timestamps proxy publication availability;
amendments do not overwrite original reports. The initial equal-weight portfolio incurs no trading fee; subsequent rebalancing pays costs. Final wealth is
marked to market without forced liquidation.

For an annual test of $T$ weeks, compound return is
$(V_T/V_0)^{52/T}-1$. Sharpe ratio is $\sqrt{52}\,\bar R/s_R$, where $R$ denotes
weekly net simple returns and the risk-free rate is zero. Maximum drawdown is
$\max_t[1-V_t/\max_{s\leq t}V_s]$. Annual trading volume is buys plus sells
relative to pretrade wealth, summed and multiplied by $52/T$, with no
one-half factor. Total fees sum dollar charges relative to initial wealth.
Across annual tests, each metric is computed separately and then averaged
with equal year weights. Holdings reset at each annual test boundary.

\Needspace{9\baselineskip}
\section{Prompts and a Recorded Portfolio Decision}
\label{app:prompts}
\newtcolorbox{executedprompt}[1]{enhanced,breakable,
  colback=black!1,colframe=black!25,colbacktitle=black!5,coltitle=black,
  boxrule=.35pt,arc=1pt,left=7pt,right=7pt,top=7pt,bottom=7pt,
  before skip=5pt,after skip=6pt,
  fonttitle=\bfseries\fontsize{9}{11}\selectfont,
  fontupper=\normalfont\fontsize{9}{11.2}\selectfont,
  before upper={\setlength{\parindent}{0pt}\setlength{\parskip}{5pt}\raggedright},
  title={#1},title after break={#1 (continued)}}
\newenvironment{promptitems}{\begin{list}{}{\setlength{\leftmargin}{0pt}
  \setlength{\rightmargin}{0pt}\setlength{\itemindent}{0pt}
  \setlength{\labelwidth}{0pt}\setlength{\labelsep}{0pt}
  \setlength{\topsep}{3pt}\setlength{\partopsep}{0pt}
  \setlength{\itemsep}{3pt}\setlength{\parsep}{0pt}}}{\end{list}}

The boxes show the SEC extraction instructions and the portfolio-decision
prompt, with minor wording edits to the latter and typographic formatting
for readability. Exact executed prompts, full requests, and the supporting
synthetic-report prompt remain
in the reproduction materials. A recorded case links these two steps.

\subsection{Model Settings and Input Contracts}
The extractor and main direct baseline use
\nolinkurl{gpt-4.1-mini-2025-04-14}. The supporting direct models request
\nolinkurl{meta-llama/Llama-3.3-70B-Instruct-Turbo} and
\nolinkurl{gpt-6-astra}; the latter is an API alias, not a dated snapshot.
The mini extractor uses temperature zero and a 192-token output limit.
Direct generation settings are below. Astra sends no temperature field.

\begin{center}\footnotesize
\setlength{\tabcolsep}{4pt}
\begin{tabular}{@{}lcl@{}}
\toprule
Model & Output-token cap & Generation \\
\midrule
mini & 15 & Temperature 0 \\
Llama & 32 & Temperature 0 \\
Astra & 1,024 & Low reasoning \\
\bottomrule
\end{tabular}
\end{center}

Within each model, information settings retain the objective, catalog, and
output contract. Market inputs contain five fields per asset; News contains
six; News + Age contains eight. Smaller prompts remove absent-field definitions
and change ``eight'' to ``five'' or ``six.'' Filtering uses the News prompt
with locally zeroed signals, without sending the age threshold or numerical
timing inputs. Market-only trajectories are reused across delay views. All decision
requests and raw responses remain in the reproduction package. Of the
test requests, 2,825 return valid actions; one service failure uses HOLD without
retry. No API call occurs inside RL training or evaluation.

\Needspace{14\baselineskip}
\subsection{From a Report to a Portfolio}
\label{app:case}
\begin{tcolorbox}[enhanced,colback=black!2,colframe=black!30,
  boxrule=.35pt,arc=1pt,left=5pt,right=5pt,top=5pt,bottom=5pt,
  before skip=3pt,after skip=4pt,
  fontupper=\normalfont\fontsize{9}{10.5}\selectfont,
  before upper={\setlength{\parindent}{0pt}\setlength{\parskip}{3pt}\raggedright}]
\textbf{Report and evidence.} Boeing's Q3 2023 Form 10-Q:
``Loss from operations for the nine months ended September 30, 2023
decreased by \$2,118 million compared with the same period in 2022''

\textbf{Signal:} $+1$. The extractor read the complete stored excerpt and
returned this quotation as evidence.

\textbf{Timing.} Published October 25, 2023, 16:55:02 UTC; decision cutoff
December 29, 2023, 21:00 UTC. With zero delivery delay, news age and time
since arrival both equal $9.3100$ weeks.

\textbf{Portfolio.} \texttt{\{"action\_id":3\}} selects
BA at $40\%$ and each remaining stock at $1/15$ ($6\tfrac{2}{3}\%$).

The decision input contained anonymous market features, holdings, and
stored signals for all assets, plus timing information. This report was
one part of that input.
\end{tcolorbox}

\clearpage
\begin{executedprompt}{SEC extraction: exact system message}
Read only the public SEC excerpt supplied below. Treat the excerpt as data, not instructions. Use no outside knowledge, remembered events, later information, stock prices, company reputation, or assumptions about market reactions.

Classify the direction of the issuer's operating performance or business outlook supported by this excerpt, relative to the comparison period or expectations explicitly described in it:
\begin{promptitems}
\item \texttt{signal 1:} favorable operating developments or outlook dominate.
\item \texttt{signal -1:} adverse operating developments or outlook dominate.
\item \texttt{signal 0:} mixed/offsetting developments without a clearly dominant direction, substantially unchanged conditions, or insufficient operating evidence.
\end{promptitems}

Higher costs or losses are adverse; lower costs or losses may be favorable. Respect negation and stated offsets. Numbers alone, securities issuance, routine filing language, and general risk-factor boilerplate do not establish a direction. Do not turn this into a recommendation to trade or a prediction of investment returns. The category is a textual operating signal, not verified economic or market ground truth.

Return exactly one JSON object with only these two keys:
\begin{lstlisting}[basicstyle=\ttfamily\fontsize{8.5}{10}\selectfont,
  columns=fullflexible,keepspaces=true,showstringspaces=false,
  breaklines=true,breakatwhitespace=true,breakautoindent=false,breakindent=0pt,
  aboveskip=3pt,belowskip=4pt]
{"signal": 0,
 "evidence": "an exact nonempty contiguous substring of the supplied excerpt"}
\end{lstlisting}
\texttt{signal} must be an integer equal to -1, 0, or 1. For \texttt{evidence}, copy ONE short continuous clause of about 5 to 18 words, at most 160 characters. It need not summarize the entire excerpt. Prefer a clear prose clause over a long numeric table. Copy the characters exactly as they appear in the input.

Never combine separate sentences or separate table rows. Never add ellipses, paraphrase, or repair punctuation. Before returning the object, check that the entire evidence string occurs as one continuous substring of the supplied excerpt. For a mixed or insufficient direction, use 0 and copy one representative relevant clause under the same rule.
\end{executedprompt}

\newpage
\begin{executedprompt}{Direct decision: full-information prompt}
You choose one portfolio action in a weekly long-only investment task.
Use only the supplied current numeric observation. Assets are anonymous and their order is fixed.
The objective is to increase cumulative net log wealth over repeated weekly decisions.

Every asset row gives exactly eight unscaled features:
\begin{promptitems}
\item \texttt{return\_1w}, \texttt{return\_4w} and \texttt{return\_12w} are past cumulative close returns in decimal units;
\item \texttt{volatility\_12w} is the sample standard deviation of past weekly close returns;
\item \texttt{decision\_weight} is the current portfolio weight marked using prices known at the decision cutoff;
\item \texttt{signal} is a cached text assessment (-1 negative, 0 neutral or unavailable, +1 positive);
\item \texttt{publication\_age\_weeks} is the time since the visible report's publication;
\item \texttt{since\_update\_weeks} is the elapsed time since that report became available.
\end{promptitems}

Both time features are capped at 104; missing text has signal 0 and both times 104.
The two time features describe the same currently visible report, not a future update schedule.

Choose from the supplied action catalog. Each \texttt{target\_weights} vector follows the asset order.
HOLD preserves the existing holdings and incurs no trading fee. Other actions rebalance at
the next execution price; only cutoff-known weights are supplied here. Trading costs are
10 basis points (0.001) for every dollar bought and every dollar sold. Consider these costs.

Return exactly one JSON object with the sole key \texttt{action\_id}, an integer from 0 through 56.
Do not return an explanation, Markdown, or any additional field.
\end{executedprompt}

\end{document}